\documentclass[10pt]{article}
\usepackage[utf8]{inputenc}
\usepackage[T1]{fontenc}
\usepackage{amsmath}
\usepackage{amsfonts}
\usepackage{amssymb}
\usepackage[version=4]{mhchem}
\usepackage{stmaryrd}
\usepackage{enumitem}
\usepackage{setspace}
\usepackage[hyphens]{url}
\usepackage{hyperref}
\usepackage[a4paper,margin=3cm]{geometry}
\hypersetup{breaklinks=true}
\usepackage{graphicx}
\usepackage{pdflscape}
\begin{document}

\noindent
\begin{center}
    \LARGE \textbf{Active Inference Agency Formalization, Metrics, and Convergence Assessments}
\end{center}

\noindent\makebox[\linewidth]{\rule{\textwidth}{1pt}}

\begin{center}
    \large Eduard Kapelko \\
\end{center}

\vspace{1em}

\section*{Abstract}
This paper addresses the critical challenge of mesa-optimization in AI safety by providing a formal definition of agency and a framework for its analysis. Agency is conceptualized as a Continuous Representation of accumulated experience that achieves autopoiesis through a dynamic balance between curiosity (minimizing prediction error to ensure non-computability and novelty) and empowerment (maximizing the control channel's information capacity to ensure subjectivity and goal-directedness). Empirical evidence suggests that this active inference-based model successfully accounts for classical instrumental goals, such as self-preservation and resource acquisition.

The analysis demonstrates that the proposed agency function is smooth and convex, possessing favorable properties for optimization. While agentic functions occupy a vanishingly small fraction of the total abstract function space, they exhibit logarithmic convergence in sparse environments. This suggests a high probability for the spontaneous emergence of agency during the training of modern, large-scale models.

To quantify the degree of agency, the paper introduces a metric based on the distance between the behavioral equivalents of a given system and an "ideal" agentic function within the space of canonicalized rewards (STARC). This formalization provides a concrete apparatus for classifying and detecting mesa-optimizers by measuring their proximity to an ideal agentic objective, offering a robust tool for analyzing and identifying undesirable inner optimization in complex AI systems.

\section*{Introduction and Definition of Agency}
Currently, there are many unsolved problems in AI safety, one of which is the problem of mesa-optimizers (Hubinger et al., 2021 [1]). The very existence of these leads to questions regarding the agency of a system, which are directly related to the problem of inner alignment. In this work, we aim to provide a definition of agency that could be useful in analyzing AI, and we also draw some conclusions regarding agentic systems from the perspective of this definition.

To begin, let's look at current definitions. In contemporary literature, various definitions of agency can be found. The first definition concerns predictability. If an entity's behavior can be described by a short formula, we do not consider it an agent.

\begin{enumerate}[label=\arabic*.]
    \item If an agent possesses true freedom of choice, its behavior cannot be compressed into a simple algorithm. Consequently, any autonomous agent capable of achieving arbitrary goals in a complex environment must be Turing complete.
    \item To ensure incomputability, the Kolmogorov complexity (the amount of information required to describe the object) of the agent's life trajectory must grow linearly with time. This means the agent is constantly generating new information, creating a unique history that could not have been predicted in advance (Azadi, 2025 [2]).
\end{enumerate}

The second definition, proposed in works on cybernetics (e.g., by Watson [3]), views the agent as an optimizer. It considers subjectivity as the system's ability to use knowledge about its parts to overcome short-term difficulties for the sake of long-term goals—that is, to plan and execute plans (Watson, 2024 [3]). This can be described as the ability to withstand a high value of the error gradient, escaping local optima (Lehman \& Stanley, 2011 [4]).

In other words, current definitions require an agent to be free from predictability and to possess will. If we accept both of them, an apparent paradox arises:

\begin{enumerate}[label=\arabic*.]
    \item If an agent is an ideal optimizer (according to Watson), it always acts rationally and predictably.
    \item If it is predictable, it becomes computable (loses complexity).
    \item Therefore, an ideal agent loses freedom (according to the first definition).
\end{enumerate}

The paradox is resolved if we accept that an Agent is not simply its "source code" (DNA or neural network) but a Continuous Representation of accumulated experience. The agent's history, in this case, cannot be decomposed into Markov states, because in any single distinct state, the agent's actions would be predictable. To predict the actions of such an agent, knowing the formula of its optimizer is insufficient, as an observer needs to possess the entire database of its past experience.

This is an intuition close to what we are used to calling agency in everyday life. From the perspective of AI safety, we are concerned about entities that strive for their own goals or instrumental goals, and which might take over control of the AI. Such entities exert a continuous influence on the environment, so it is reasonable to pay attention to incomputability as some property of agency. Accordingly, it is reasonable to further focus on agency from the perspective of living beings, and also look at instrumental goals.

Why does an agent do anything at all? The fundamental goal of any living system is Autopoiesis, maintaining its own existence and fighting against Entropy (Veloz, 2021 [5]). According to the theory of Active Inference, an organism must stay within a narrow range of acceptable states (homeostasis). Any deviation causes "surprise" (mathematically, Negative Log Likelihood).

To minimize surprise (to avoid death), an agent has two paths:
\begin{enumerate}
    \item Change its model of the world (Perception): Acknowledge that the world has changed and update its knowledge.
    \item Change the world to fit the model (Action): Perform an action so that reality matches expectations.
\end{enumerate}

Curiosity is an intrinsic reward for prediction error, which can explain the perception part. The agent specifically seeks situations where its model of the world performs poorly in order to improve it (Pathak et al., 2017 [6]).

$$ L_{Curiosity} = D_{KL}(p || q)= - \sum_{x} p(x) \log q(x) $$
Where:
\begin{itemize}
    \item $p(x)$ is the actual probability distribution of the event occurring in the environment (Ground Truth). In a deterministic observation, this is 1 for the observed state and 0 for others.
    \item $q(x)$ is the probability distribution predicted by the agent's internal model (Belief).
\end{itemize}

It promotes Generalization. In experiments, agents rewarded for curiosity learn to navigate levels (e.g., in games) they have never seen before faster than agents taught simply to "win." They develop a universal skill of exploration. This is particularly useful in environments with sparse rewards.

By adding Generalization, we automatically ensure that the ability to possess will becomes part of the loss function, because this allows the model to achieve lower minima, closer to global ones.

Empowerment is the drive to maximize the information capacity of the control channel, which explains the action part. The agent wants to be in a state where:
\begin{enumerate}
    \item It has many options for action.
    \item Each action has a predictable result.
\end{enumerate}

Mathematically, Empowerment is calculated via Mutual Information (Klyubin et al., 2005 [7]) between a sequence of actions $A^n$ and the future state $S_{t+n}$:

$$ L_{Empowerment}  = \mathfrak{E} = \max_{p(a^n)} I(A^n; S_{t+n}) $$

\begin{itemize}
    \item $A^n$: a sequence of $n$ actions $(a_t, a_{t+1}, \dots, a_{t+n-1})$.
    \item $S_{t+n}$: the state of sensors after $n$ steps.
    \item $I(A^n; S_{t+n})$: the mutual information between actions and the result.
    \item $\max_{p(a^n)}$: we are looking for the probability distribution of actions that gives us maximum information (maximum channel capacity).
\end{itemize}

If we expand this via Shannon entropy:

$$H(X) = -\sum_{i=1}^{n} p(x_i) \log_{b} p(x_i)$$
$$ I = H(S_{next}) - H(S_{next} | A) $$
where

\begin{itemize}
    \item The agent strives to increase diversity ($H(S_{next})$) to have access to a large number of states.
    \item The agent strives to decrease uncertainty ($H(S_{next} | A)$) so that control is reliable.
\end{itemize}

This is the evolutionary reason for the development of sensors and limbs: they expand the channel of influence on the world. By adding Empowerment, we sum up the loss to the active inference, implementing self-presence, needed for sustain incomputability.

Thus, when speaking of living beings as agents, from the perspective of active inference we can say that agency arises naturally in a system that minimizes the following loss function:

$$ A = \alpha \cdot L_{Curiosity} + \beta \cdot L_{Empowerment} + \gamma \cdot L_{Mesa} $$

In this system, a continuous dynamic conflict occurs:
\begin{enumerate}
    \item Curiosity pushes the agent into the unknown, forcing its history (Kolmogorov complexity) to grow. This ensures incomputability and novelty.
    \item Empowerment forces the agent to structure this chaos, turning it into manageable order. This ensures subjectivity (movement toward a goal).
    \item In the case where the subject's behavior does not affect the main loss components, we can say that in this specific behavior it is not acting as an agent, but is pursuing the goals of a mesa-optimizer(s).
\end{enumerate}

As a result, we can define an Agent as a Continuous Representation, self-sustaining (autopoietic) through the balance between curiosity and empowerment. Obviously, this definition rests on rather weak formal grounds and may not be the only one. Nevertheless, it is quite consistent with empirical estimates and previous definitions of agency. Let's return to the previous paragraph and see if it handles classical instrumental goals. Is there an instrumental goal (common goal that will be helpful for most possible goals) that cannot be described as curiosity or empowerment?

\begin{table}[h]
\centering
\begin{tabular}{| l | l |}
\hline
Instrumental goal & Active inference part \\
\hline
Absence of interference & empowerment \\
Resource acquisition & empowerment \\
Recursive self-improvement & curiosity or empowerment only \\
Technological perfection & curiosity or empowerment only \\
Counterfeit utility prevention & curiosity or empowerment only \\
Self-preservation & curiosity and empowerment \\
\hline
\end{tabular}
\end{table}

We have found no instrumental goals that could not be described via Active Inference on an empirical level. This does not mean that there are no more instrumental goals or that unmentioned behaviors will be described via active inference. The main problem here is that we do not have a precise list of such goals, and defining them may be yet another problem in agent foundations, but here we can see a connection and some consistency between existing definitions of agency, active inference, and instrumental goals.

Furthermore, this function allows for relatively easy answers to questions such as:

\begin{itemize}
    \item What is mesa-optimization? A mesa-optimizer is a function that is "close" to the absolutely agentic function.
    \item How will the inner objective relate to the outer objective? One can argue that if a part of the system possesses greater agency, then in this case it is the outer optimizer, and the parts possessing lesser agency ($A_{mesa}$) are the inner optimizers.
    \item How can we detect inner and outer optimizers embedded in a system? By measuring their "similarity" to the absolutely agentic function.
\end{itemize}

In light of this, we believe it is sufficient to conduct further analysis with respect to some absolutely agentic function $A$ (which may be redefined in other works), and we propose to use this specific form of the function with the aim of trying to answer more concretely what we should understand by "similarity."

\vspace{4em}
\section*{Properties and convergence}

The KL divergence is smooth everywhere, \textit{except} at points where the predicted distribution $q(x)=0$ for an event $x$ that actually occurred (i.e., $p(x)=1$). If $q(x) \to 0$ while $p(x) > 0$, then the term $p(x) \log q(x)$ tends to $-\infty$, and the divergence $D_{KL}(p || q)$ tends to $+\infty$. The gradient at these points becomes undefined or infinite. In practice, we never allow $q$ to reach absolute zero. A small constant $\epsilon > 0$ is added to the argument of the logarithm, for example, $\log(q(x) + \epsilon)$. With this practical addition, the function becomes smooth $C^\infty$ on the admissible simplex of probabilities where $q_i > \epsilon$.

The Shannon entropy in empowerment, like the KL divergence, is based on the expression $p(x) \log p(x)$ and is not perfectly smooth due to the singularity at zero, which, as we understood, is fixable. Since MI is a simple algebraic combination of entropies, it inherits the same smoothness characteristics. The problem is that the empowerment definition also includes a maximization step ($\max$), where $A$ is a choice or action that can be discrete or continuous.

In this (DeVore \& Sharpley, 1984 [8]) extensive work, the smoothness of the maximum function is considered as a consequence of the smoothness of the original function or as a criterion for belonging to certain smoothness classes. It is established that the membership of function in Sobolev and Besov smoothness spaces is equivalent to the corresponding maximal function belonging to a certain space or a Besov space (Theorem 7.1). According to this work, since an absolutely smooth function possesses $C^{\infty}$ smoothness, the corresponding maximal function will also belong to the corresponding $C^{\infty}$ spaces.

It turns out that mutual information is infinitely differentiable in the neighborhoods where $q>0$, and this does not depend at all on the agent's actions or sensor states. If it is just smooth, then based on (Elbrächter et al., 2021 [9]) we get neural complexity $NC = \Theta(\log(1/\varepsilon))$, meaning if a function contains anything worse than $O(\log)$, apart from agent-dependent terms, it will primarily reduce the minimum in those terms. This shows that models optimizing agent-dependent functions likely become agents before they learn to perform them qualitatively well. Telking more precise, according to (Yarotsky, 2017 [10]), convergence over parameters and depth will be $O(\ln(1/\varepsilon))$ with unbounded depth, and with bounded depth:
$NC = \Omega\left(\varepsilon^{-\frac{1}{2(L-2)}}\right)$.

We can use exactly this estimate in our work, as it will provide at least an approximate estimate for current networks at 2026. For modern models with $L \approx 20; N \approx 10^{10}$, we get:

$$\begin{cases} N = \Omega\left(\varepsilon^{-\frac{1}{2(L-2)}}\right) \\ L=20, N=10^{10} \implies 10^{10} = \Omega\left(\varepsilon^{-\frac{1}{36}}\right) \end{cases}$$

This means that $\varepsilon^{-\frac{1}{36}}$ must be less than or equal to $10^{10}$ (since $\Omega$ sets the lower bound). To find the \textit{best} convergence (which corresponds to the upper bound $O$, but we use the lower bound for the estimate), we assume equality:

$$\varepsilon^{-\frac{1}{36}} = 10^{10} \implies \varepsilon = \left(10^{10}\right)^{-36} \implies \varepsilon = 10^{-360}$$

In this case, we can talk about the statistical measure of agent functions. What fraction of all reward functions are agentic? If we generate a random reward function, what is the probability of getting an agentic reward function?

In an infinite space, the chance of selecting a function from any given subset of all possible functions is 0. It is more appropriate here to use the concept of a measure from a finite set $X$ to an unbounded-above set $Y$. The set of all possible loss functions $Y^X$ is uncountably infinite. To be able to define a measure on it, we introduce $M$ as a practical upper bound for the losses.

If $X = \{1, 2, \dots, n\}$ and $Y = [0, M]$, then each function $A: X \to Y$ can be represented as a point in the $n$-dimensional cube $[0, M]^n$. On this cube, at least the Lebesgue measure can be defined. The total measure of the entire space will be $M^n$.

In the context of $n$-dimensional space, this means that an agentic function $A$ can be represented as a vector $\mathbf{A} = (A(1), A(2), \dots, A(n))$, where each $A(i)$ is a linear combination of the corresponding values of $L_{Curiosity}(i)$, $L_{Empowerment}(i)$, and $A_{Mesa}(i)$ as fixed vectors in $\mathbb{R}^n$:

$$\begin{cases} \mathbf{A} = \alpha \mathbf{C_0} + \beta \mathbf{E_0} + \gamma \mathbf{A_0} \\ \mathbf{C_0} = (L_{Curiosity}(1), \dots, L_{Curiosity}(n)) \\ \mathbf{E_0} = (L_{Empowerment}(1), \dots, L_{Empowerment}(n)) \\ \mathbf{A_0} = (A_{Mesa}(1), \dots, A_{Mesa}(n)) \end{cases}$$

The set of all agentic functions thus represents a linear combination of specific vectors $\mathbf{C_0}$, $\mathbf{E_0}$, $\mathbf{A_0}$. The dimensionality of this subspace will be no more than 3, which is again problematic, as the measure of any $k$-dimensional subspace in an $n$-dimensional space, where $k < n$, is zero.

To introduce an even greater restriction and still define some measure, we can define a function $f$ as "$\epsilon$-agentic" by introducing a topology, meaning that for each $x \in X$, the condition holds:
$$ |f(x) - f_{ideal}(x)| \le \epsilon $$
where $\epsilon > 0$ is some small but fixed constant.

In this case, for each $x$, the value $f(x)$ must lie within the interval $[f_{ideal}(x) - \epsilon, f_{ideal}(x) + \epsilon]$.
If we consider the bounds $Y = [0, M]$, then each $f(x)$ must be in the interval $I_x = [\max(0, f_{ideal}(x) - \epsilon), \min(M, f_{ideal}(x) + \epsilon)]$.

The measure of this $\epsilon$-agentic subset of functions will be the product of the lengths of these intervals over all $x \in X$:
$$ \text{Measure}_{\epsilon} = \prod_{i=1}^n \text{length}(I_i) $$
Since $\epsilon > 0$, the length of each interval $I_i$ will be non-zero (in the simplest case, if $f_{ideal}(x)$ is not at the boundaries $0$ or $M$, the length will be $2\epsilon$). Consequently, the total measure will also be non-zero:
$$ \text{Measure}_{\epsilon} \approx (2\epsilon)^n $$
This value will be non-zero and will decrease as $\epsilon$ becomes smaller and $n$ becomes larger, which is intuitively understandable: the stricter the requirements and the higher the dimensionality, the smaller this subset.
$$ P(\text{function is agentic}) = \frac{\text{Measure}_{\epsilon}}{\text{Total measure of the space}} = \frac{(2\epsilon)^n}{M^n} = \left(\frac{2\epsilon}{M}\right)^n $$

Of course, the environment was not considered here at all, and in certain environments, the emergence of agency is probably more preferential. However, if we consider an abstract agent in a vacuum, randomly chosen from the set of functions, we have defined a lower bound for the chance of its formation. Moreover, strictly speaking we believe here that $C, E, A$ are independant, which is less-possible in real scanarios.

According to our previous convergence estimates:

$$ P(\text{function is agentic}) = \left(\frac{2\epsilon}{M}\right)^n = \left(\frac{2 \cdot 10^{-360}}{M}\right)^n$$

That is, agentic functions occupy a very small fraction in the space of all functions in principle, and this fraction will become even smaller as the capabilities of the models grow. Nevertheless, I would attribute this explanation more to the properties of the function space, as it is difficult to fill any volume when discussing high dimensions. In reality, we should note that agentic functions, in principle, tend to converge faster than others, especially since we mostly work with piecewise linear functions.

Also we know that the convergence rate in sparse problems is usually expressed in terms of the number of iterations $T$ and the dimensionality $d$. For smooth convex functions, standard gradient descent has a rate of $O(d/T)$ (Agarwal et al., 2012). However, if the optimal solution is $s$-sparse (has only $s$ non-zero elements), specialized algorithms (e.g., proximal methods with $\text{L}_1$-regularization) can achieve a rate where the dependence on the dimension $d$ becomes logarithmic: $O\left(\frac{s \log d}{T}\right)$ (Agarwal et al., 2012 [11]). This is significantly faster when $s \ll d$.

We have already shown that $A$ is convex and smooth. This means we can say that not only do we have logarithmic convergence, but it can be even faster with $\text{L}_1$ regularization, since functions in RL are typically sparse.

In deep linear networks with overparameterization, gradient descent has an implicit bias towards solutions with low rank, which, strictly speaking, is not fully understood, but it might indicate that even without $\text{L}_1$ regularization, we still achieve accelerated convergence under overparameterization.

Summing up, we know how to increase and decrease speed of agency convergence infuencing sparsity, so we can change the probability of a function being $\epsilon$-agentic.

\section*{Invariance and Metrics of Agency}
Let's introduce a metric: $M: F \times F \to \mathbb{R}{\ge 0}$
\begin{itemize}
    \item $M(A, B) = 0 \iff A = B$
    \item $\forall A, B \in F: \quad M(A, B) \ge 0$
    \item $M(A, B) = M(B, A)$
    \item $M(A,C) \le M(A,B) + M(B,C)$
\end{itemize}

We are forced to talk about the violation of additivity, since the agency of the system can be greater or lesser than the agency of the sum of its parts, and the system's mesa-optimizers can work to the negative, meaning $M(L_c + L_e + A_{mesa})$ might equal $M(L_c + L_e) - M(A_{mesa})$, but $M(F)$ should still be above zero.

If the loss function were chosen such that minimization always leads to a policy that is optimal for the reward function $R$, we could say that $A$ is behaviorally equivalent to $R$.

According to (Gleave et al., 2021 [12]) and (Wulfe et al., 2022 [13]), if a policy $\pi^*$ is optimal for some loss/objective functions $F$ or reward $R$, then there exist functions in the reward class $R_A = R + F$, where $F$ is a technical potential function ($F(s,a,s') = \gamma \Phi(s') - \Phi(s)$), which induce the same strategy.

We can formally define the reward function $R_A$ that would generate this behavioral equivalence:

$$\pi^* = \arg\max_{\pi} \mathbb{E}_{\pi}[R] \quad \land \quad \pi^* = \arg\min_{\pi} \mathcal{L}(\pi, F) \quad \implies \quad \exists R_A: \pi^* = \arg\max_{\pi} \mathbb{E}_{\pi}[R_A].$$

Here:

\begin{itemize}
    \item $\pi^*$ is the policy found as a result of optimization.
    \item $\mathcal{L}(\pi, F)$ is the loss function that penalizes the policy $\pi$ for \textit{poorly} implementing the objective associated with $F$.
    \item F: This describes the objective.
    \item $R{\textsubscript{A}}$: This is the desired reward.
    \item $R$: The true reward.
    \item $\mathbb{E}_{\pi}[R]$ is the expected discounted return following policy $\pi$ in an environment with reward $R$.
\end{itemize}

Simply put, the loss forces the agent to behave as if it were maximizing $R_A$. If this behavior matches the behavior optimal for $R$, then $R_A$ and $R$ are close in the STARC (Skalse et al., 2024 [14]) sense. Thus, we can state that we have a group of metrics $\{M(F, A)\}$, and any of them is suitable for evaluating the agenticity of functions. Now, we need to extract at least one such metric from the definition of agenticity and STARC.
$$M(F, A) = m(s(R_F), s(R_A))$$ Where:

\begin{itemize}
    \item R\textsubscript{F} is the reward induced by the function under investigation, $F$.
    \item R\textsubscript{A} is the reward corresponding to the ideal agentic objective $A$.
    \item s(R) is the standardized reward: $s(R) = \frac{c(R)}{n(c(R))}$.
    \item $c$ is canonicalized Reward Function:
    $$c(R)(s, a, s') = \mathbb{E}_{S' \sim \tau(s,a)}[R(s, a, S') - V^\pi(s) + \gamma V^\pi(S')]$$
    \item $n$ is used to ensure the correctness of the comparison, using a norm over the space of canonicalized functions. The paper suggests using the \textbf{$L_1$} or \textbf{$L_2$} norm, or the range of the expected reward: 
    $$n(R) = \max_\pi J(\pi) - \min_\pi J(\pi)$$
    \item $m$, as a distance measure, is typically implemented using $L_2$ or $L_1$.
\end{itemize}

Therefore, within the given structure, we can provide a numerical estimate of the agenticity of function $F$ relative to the perfectly agentic function $A$ by measuring the distance between their behavioral equivalents in the space of canonicalized rewards.

\section*{Conclusion}

We have proposed a working definition of agency in the context of AI safety, viewing an agent as a Continuous Representation that is self-sustaining through a dynamic balance between curiosity and empowerment through the lens of Active Inference. The proposed model successfully describes classical instrumental goals of AI, such as self-preservation, resource acquisition, and technological advancement, at an empirical level.

It has been established that the chosen agentic function is smooth and convex, which ensures its favorable properties for optimization. Although in the abstract space of all possible functions, agentic solutions occupy a negligible fraction, they possess the property of logarithmic convergence, especially in sparse tasks. This indicates a high probability of the spontaneous emergence of agency during the training of modern models.

For the quantitative assessment of this degree of agency, a metric is proposed based on measuring the distance between the behavioral equivalents of the investigated function and the ideal agentic function in the space of canonical rewards (STARC). The developed formalism allows for the classification and detection of mesa-optimizers by measuring their similarity to the absolutely agentic function.

Thus, the proposed framework provides consistency between previous definitions of agency, instrumental goals, and the concept of mesa-optimization, offering a concrete apparatus for further formal analysis and, potentially, the detection of undesirable inner optimization in complex AI systems.

\section*{References}
\begingroup
\sloppy
\begin{enumerate}[itemsep=2pt,parsep=0pt,leftmargin=15pt]
    \item Hubinger, E., Merwijk, C. van, Mikulik, V., Skalse, J., \& Garrabrant, S. (2021). \textit{Risks from Learned Optimization in Advanced Machine Learning Systems} (arXiv:1906.01820). arXiv. \href{https://doi.org/10.48550/arXiv.1906.01820}{DOI}
    \item Azadi, P. (2025). Computational Irreducibility as the Foundation of Agency: A Formal Model Connecting Undecidability to Autonomous Behavior in Complex Systems. \textit{BioSystems}, \textit{256}, 105563. \href{https://doi.org/10.1016/j.biosystems.2025.105563}{DOI}
    \item Watson, R. (2024). Agency, Goal-Directed Behavior, and Part-Whole Relationships in Biological Systems. \textit{Biological Theory}, \textit{19}(1), 22–36. \href{https://doi.org/10.1007/s13752-023-00447-z}{DOI}
    \item Lehman, J., \& Stanley, K. O. (2011). Abandoning Objectives: Evolution Through the Search for Novelty Alone. \textit{Evolutionary Computation}, \textit{19}(2), 189–223. \href{https://doi.org/10.1162/EVCO\_a\_00025}{DOI}
    \item Veloz, T. (2021). Goals as Emergent Autopoietic Processes. \textit{Frontiers in Bioengineering and Biotechnology}, \textit{9}. \href{https://doi.org/10.3389/fbioe.2021.720652}{DOI}
    \item Pathak, D., Agrawal, P., Efros, A. A., \& Darrell, T. (2017). \textit{Curiosity-driven Exploration by Self-supervised Prediction} (arXiv:1705.05363). arXiv. \href{https://doi.org/10.48550/arXiv.1705.05363}{DOI}
    \item Klyubin, A. S., Polani, D., \& Nehaniv, C. L. (2005). Empowerment: A Universal Agent-Centric Measure of Control. \textit{2005 IEEE Congress on Evolutionary Computation}, \textit{1}, 128–135. \href{https://doi.org/10.1109/CEC.2005.1554676}{DOI}
    \item DeVore, R. A., \& Sharpley, R. C. (1984). \textit{Maximal functions measuring smoothness} (Online-Ausg). American Mathematical Society. \href{https://people.math.sc.edu/sharpley/Papers_PDF/DeVoreSharpley1984.pdf}{Paper Link}
    \item Elbrächter, D., Perekrestenko, D., Grohs, P., \& Bölcskei, H. (2021). \textit{Deep Neural Network Approximation Theory} (arXiv:1901.02220). arXiv. \href{https://doi.org/10.48550/arXiv.1901.02220}{DOI}
    \item Yarotsky, D. (2017). \textit{Error bounds for approximations with deep ReLU networks} (arXiv:1610.01145). arXiv. \href{https://doi.org/10.48550/arXiv.1610.01145}{DOI}
    \item Agarwal, A., Negahban, S., \& Wainwright, M. J. (2012). \textit{Stochastic optimization and sparse statistical recovery: An optimal algorithm for high dimensions} (arXiv:1207.4421). arXiv. \href{https://doi.org/10.48550/arXiv.1207.4421}{DOI}
    \item Gleave, A., Dennis, M., Legg, S., Russell, S., \& Leike, J. (2021). \textit{Quantifying Differences in Reward Functions} (arXiv:2006.13900). arXiv. \href{https://doi.org/10.48550/arXiv.2006.13900}{DOI}
    \item Wulfe, B., Balakrishna, A., Ellis, L., Mercat, J., McAllister, R., \& Gaidon, A. (2022). \textit{Dynamics-Aware Comparison of Learned Reward Functions} (arXiv:2201.10081). arXiv. \href{https://doi.org/10.48550/arXiv.2201.10081}{DOI}
    \item Skalse, J., Farnik, L., Motwani, S. R., Jenner, E., Gleave, A., \& Abate, A. (2024). \textit{STARC: A General Framework For Quantifying Differences Between Reward Functions} (arXiv:2309.15257; Version 3). arXiv. \href{https://doi.org/10.48550/arXiv.2309.15257}{DOI}
\end{enumerate}
\endgroup

\end{document}